\documentclass[10pt,twocolumn,letterpaper]{article}

\usepackage{cvpr}              % To produce the CAMERA-READY version
\usepackage{graphicx}
\usepackage{booktabs}
\usepackage{makecell}
\usepackage{multirow}
\usepackage{amsmath}
\usepackage{amssymb}
\usepackage{amsmath,amsfonts,bm}

\def\eqref#1{equation~\ref{#1}}
\def\1{\bm{1}}

\DeclareMathAlphabet{\mathsfit}{\encodingdefault}{\sfdefault}{m}{sl}
\SetMathAlphabet{\mathsfit}{bold}{\encodingdefault}{\sfdefault}{bx}{n}

\newcommand{\E}{\mathbb{E}}

\newtheorem{theorem}{Theorem} % Theorem
\newtheorem{lemma}[theorem]{Lemma} % Lemma
\usepackage{bm}
\usepackage{romannum} % roman number
\usepackage{algorithm}
\usepackage{algpseudocode}

\usepackage[pagebackref,breaklinks,colorlinks]{hyperref}

\usepackage[capitalize]{cleveref}
\crefname{section}{Sec.}{Secs.}
\Crefname{section}{Section}{Sections}
\Crefname{table}{Table}{Tables}
\crefname{table}{Tab.}{Tabs.}

\def\cvprPaperID{3931} % *** Enter the CVPR Paper ID here
\def\confName{CVPR}
\def\confYear{2022}

\begin{document}
\pagenumbering{arabic} % page is arabic number

%%%%%%%%% TITLE START
% Title candidate 1
% \title{PAD: Pruning Adversarial Vulnerability for Robustness and Compactness}

% Title candidate 2
\title{Masking Adversarial Damage: Finding Adversarial Saliency\\for Robust and Sparse Network}
%%%%%%%%% TITLE END

%%%%%%%%% Author information START
\author{Byung-Kwan Lee\thanks{Equal contribution. $\dagger$ Corresponding author.},~~Junho Kim\footnote[1]{},~~Yong Man Ro\footnote[2]{}\\
% Image and Video Systems Laboratory\\
% School of Electrical Engineering\\
% Korea Advanced Institute of Science and Technology (KAIST)\\
Image and Video Systems Lab, School of Electrical Engineering, KAIST, South Korea\\
{\tt\small \{leebk, arkimjh, ymro\}@kaist.ac.kr}
% \and
}
%%%%%%%%% Author information END
\maketitle

%%%%%%%%% ABSTRACT START
\begin{abstract}
Adversarial examples provoke weak reliability and potential security issues in deep neural networks. Although adversarial training has been widely studied to improve adversarial robustness, it works in an over-parameterized regime and requires high computations and large memory budgets. To bridge adversarial robustness and model compression, we propose a novel adversarial pruning method, Masking Adversarial Damage (MAD) that employs second-order information of adversarial loss. By using it, we can accurately estimate adversarial saliency for model parameters and determine which parameters can be pruned without weakening adversarial robustness. Furthermore, we reveal that model parameters of initial layer are highly sensitive to the adversarial examples and show that compressed feature representation retains semantic information for the target objects. Through extensive experiments on three public datasets, we demonstrate that MAD effectively prunes adversarially trained networks without loosing adversarial robustness and shows better performance than previous adversarial pruning methods.

\end{abstract}
%------------------------------------------------------------------------
\section{Introduction}
\label{sec:Introduction}

Deep neural networks (DNNs) have achieved impressive performances in a wide variety of computer vision tasks (\eg, image classification~\cite{NIPS2012_c399862d, 7780459}, object detection~\cite{NIPS2015_14bfa6bb, lin2017feature}, and semantic segmentation~\cite{he2017mask, chen2018encoder}). Despite the breakthrough outcomes, DNNs are easily deceived from adversarial attacks with carefully crafted perturbations~\cite{42503, 43405, madry2018towards, CW}. Injecting these perturbations into benign images generates adversarial examples which hinder the decision-making process of DNNs. Although the perturbations are imperceptible to humans, they easily induce vulnerable features in DNNs~\cite{NEURIPS2019_e2c420d9, kim2021distilling}. Due to such fragility, various deep learning applications have suffered from potential security issues that induce weak reliability of DNNs~\cite{8756865, WANG201912, 8824956}.

Accordingly, to achieve robust and reliable DNNs, a lot of adversarial research have been dedicated to presenting powerful adversarial attack and defense algorithms in the sense of cat-and-mouse games. Among various methods, adversarial training (AT)~\cite{45816, 43405, madry2018towards, AthalyeC018} has been widely studied to improve adversarial robustness so far, where DNNs are trained with adversarial examples. Madry~\etal~\cite{madry2018towards} have shown that adversarially trained models are robust against several white-box attacks with the knowledge of model parameters. Besides, recent studies~\cite{pmlr-v97-zhang19p, Wang2020Improving} have further enhanced the robustness by adding various regularizations to fully utilize not only the adversarial examples but also benign examples for model generalization.

Orthogonal to the adversarial issue, most of the AT-based methods work in an over-parameterized regime to achieve the robustness, thus they induce higher computations and require larger memory budgets than benign classifiers~\cite{madry2018towards}. Thereby, applying AT-based methods to resource-constrained devices is burdensome and known as a critical limitation. To bridge adversarial robustness and model compression, several studies~\cite{sehwag2019towards, MLSYS2019_ec5decca, vemparala2021adversarial, ozdenizci2021training} have introduced adversarial pruning mechanisms to reduce its model capacity while preserving the adversarial robustness.

In standard training procedure, a promising pruning method is to remove the lowest weight magnitudes (LWM)~\cite{han2015learning, han2015deep}, assuming that small magnitudes affect the least changes of the model prediction. Under its assumption, Sehwang~\etal~\cite{sehwag2019towards} have proposed a 3-step pruning method (pre-training, pruning, and fine-tuning) for the adversarial examples. Several works~\cite{ye2019adversarial,gui2019model} further have enhanced LWM pruning methods by employing alternating direction method of multipliers (ADMM)~\cite{zhang2018systematic} or Beta-Bernoulli dropout~\cite{madaan2020adversarial} to eliminate unuseful model parameters. Recently, Sehwag~\etal~\cite{sehwag2020hydra} have argued that despite successful results of LWM pruning, such heuristic pruning methods cause performance degradation when integrated with adversarial training. Accordingly, they have formulated a way of finding importance scores~\cite{ramanujan2020s} and removed model parameters with low importance scores for adversarial training loss. We further focus on how to reflect correlations of each model parameter to prune multiple parameters at once and theoretically formulate local geometry of adversarial loss to identify which combinatorial model parameters affect model prediction in adversarial settings.

In this paper, we present a novel adversarial pruning method, namely Masking Adversarial Damage (MAD) that uses second-order information (Hessian) of the adversarial loss with masking model parameters. Motivated by Optimal Brain Damage (OBD)~\cite{lecun1990optimal} and Optimal Brain Surgeon (OBS)~\cite{hassibi1993optimal} employing Hessian of the loss function, we devise a way of estimating \textit{adversarial saliency} for model parameters by considering their correlated conjunctions, which can more precisely represent the connectivity of model parameters for the adversarial prediction.

In recent works~\cite{wang2019eigendamage,singh2020woodfisher}, approximating Hessian of multiple parameters is regarded as computationally prohibitive. Alternatively, computing each parameter's importance and sorting it with pruning statistics is a practical approach in standard pruning. Bringing in such combinatorial problem into adversarial settings, we aim to prune adversarially less salient parameters that cannot arouse adversarial vulnerability. To that end, we first optimize masks for DNNs to be predictive to adversarial examples and approximate Hessian of the adversarial loss by utilizing optimized masks. To effectively compute Hessian, we introduce a block-wise K-FAC approximation that can consider the local geometry of the adversarial loss at multiple parameters points. Based on the change of the adversarial loss along with mask-applied parameters, we can track how sensitively specific conjunctions of model parameters respond to adversarial examples and compress DNNs without weakening the robustness.

For the proposed method, we thoroughly analyze adversarial saliency and compressed feature representation, and we reveal that: \romannum{1}) model parameters of initial layer are highly sensitive to adversarial perturbation, \romannum{2}) MAD can retain semantic information of target objects even with the high pruning ratio. Through extensive experiments on MAD in three major datasets, we corroborate that MAD can effectively compress DNNs without losing adversarial robustness and show better adversarial defense performance than the previous adversarial pruning methods. 

The major contributions of this paper are as follows:

\begin{itemize}
\item We present a novel adversarial pruning method, Masking Adversarial Damage (MAD) which can precisely estimate adversarial saliency for model parameters using second-order information. 

\item By analyzing MAD, we investigate that the model parameters of initial layer are highly sensitive to adversarial examples, and compressed feature representation retains semantic information of target objects.

\item Through extensive experiments, we demonstrate the effectiveness of compression capability as well as the adversarial robustness for the proposed method.

\end{itemize}

%------------------------------------------------------------------------
\section{Background and Related Work}
\label{sec:Related Work}

In this section, we specify the notations used in our paper and summarize the related works on adversarial training and model compression.

{\textbf{Notations.}} Let $x$ denote a benign image and $y$ denote a target label corresponding to the input image. Let $\mathcal{D}$ indicate a dataset such that $(x, y) \sim \mathcal{D}$. A deep neural network $f$ parameterized by model parameters $w$ is denoted by $f_{w}$. An adversarial example is represented by $x_{\text{adv}}=x+\delta^{*}$, where $\delta^{*}$ indicates adversarial perturbation as follows:
\begin{equation}
\label{eqn:attack}
\delta^{*} = \arg\max\limits_{\left \| \delta \right \|_{\infty} \leq \gamma}\mathcal{L}(f_{w}(x+\delta),y),
\end{equation}
where $\mathcal{L}$ denotes a pre-defined loss function. In this paper, we use the cross-entropy loss for image classification. We regard $\delta$ as $l_{\infty}$ perturbation within $\gamma$-ball (\ie, perturbation budget) to be imperceptible to humans such that $\left \| \delta \right \|_{\infty} \leq \gamma$. Here, $\left \| \cdot \right \|_{\infty}$ describes $l_{\infty}$ perturbation magnitude.

\subsection{Adversarial Training}
Adversarial training~\cite{43405, 45816, madry2018towards} becomes one of the intensive robust optimization methods in deep learning field, which allows DNNs to learn robust parameters against adversarial examples. In earlier work, Szegedy \etal~\cite{42503} have found that the specific nature of the adversarial examples is not a random artifact of deep learning. Then, Goodfellow \etal~\cite{43405} have introduced a single-step attack of Fast Gradient Sign Method (FGSM) that efficiently generates adversarial examples using first gradient of loss on back propagation and trained DNNs with FGSM-based adversarial examples to have an additional regularization benefit.

Madry \etal~\cite{madry2018towards} have studied adversarial robustness in a model through the lens of robust optimization (min-max game). Then, they have modified empirical risk minimization (ERM) to incorporate adversarial perturbations and proposed an adversarial training with a multi-step attack of Projected Gradient Descent (PGD). The formulation of the min-max game for the robust optimization under $l_{\infty}$ can be written as:
\begin{equation}
\label{eqn:defense}
    w^{*} = \arg\min\limits_{w} \E_{(x, y)\sim\mathcal{D}}\left[ \max\limits_{\left \| \delta \right \|_{\infty} \leq \gamma}  \mathcal{L} \left(f_{w}(x+\delta),y\right) \right].
\end{equation}
They draw a universal first-order attack through the inner maximization problem while outer minimization iteratively represents training DNNs with adversarial examples created at inner maximization one. Through it, they have designed a reliable method of adversarial training with PGD and empirically demonstrated that the acquired model parameters $w^{*}$ provide the robustness against a first-order adversary as a natural security guarantee. It allows us to cast both attacks and defenses into a common theoretical framework.

\subsection{Model Compression}

Since DNNs generally have intrinsic over-parameterized properties, they induce high computation and memory inefficiency to apply on resource-limited devices~\cite{han2015deep}. Therefore, several works~\cite{lecun1990optimal, wen2016learning, he2017channel} have proposed to prune weight parameters based on the lowest weight magnitudes (LWM). Besides, various approaches have been introduced to improve model compression utilizing absolute weight value~\cite{han2015learning}, second-order information~\cite{hassibi1993optimal,dong2017learning}, and alternating direction method of multipliers (ADMM) optimization~\cite{zhang2018systematic}. More recently, to estimate enhanced loss change when removing a single parameter, several works based on OBD~\cite{lecun1990optimal} and OBS~\cite{hassibi1993optimal} have utilized layer-wise pruning framework (L-OBS)~\cite{dong2017learning}, K-FAC approximation~\cite{wang2019eigendamage}, and Woodbury formulation (WoodFisher)~\cite{singh2020woodfisher} to effectively compute (inverse) Hessian matrix.

In parallel with various model compression methods for benign classifiers, achieving compression in adversarial settings also has drawn attention to mitigating expensive computational costs that upholds adversarial robustness of DNNs. Several works~\cite{guo2018sparse,wang2018adversarial} have investigated the relationship between the adversarial robustness and model compression and theoretically concluded that moderately compressed model may improve adversarial robustness.

Sehwang~\etal~\cite{sehwag2019towards} have demonstrated that LWM pruning can increase adversarial robustness for the adversarial examples. After that, LWM-based methods in adversarial settings become a typical pruning approach. Ye~\etal~\cite{ye2019adversarial} have improved LWM pruning method by employing ADMM optimization. Gui~\etal~\cite{gui2019model} have introduced a unified framework called ATMC including pruning, factorization, and quantization using ADMM. Recently, Sehwang~\etal~\cite{sehwag2020hydra} have argued that such heuristic LWM-based methods provoke performance degradation and proposed a generalized formulation of computing weight importance for adversarial loss using importance score optimization~\cite{ramanujan2020s}. Madaan~\etal~\cite{madaan2020adversarial} have defined adversarial vulnerability in feature space and integrated it into adversarial training to suppress feature-level distortions. In this paper, we propose a method for estimating adversarial saliency using second-order information, which serves as a key to prune parameters without weakening adversarial robustness.

\section{Methodology}
\label{sec:Method}

\subsection{Revisiting Second-order Information Pruning}
\label{sec:obd}
{\textbf{Optimal Brain Damage (OBD). }}For well-trained DNNs, Lecun \etal~\cite{lecun1990optimal} have removed a single model parameter and computed a loss change obtained from the removed single parameter. Their goal in pruning is to remove parameters that do not significantly change the loss function. Thus, they can selectively delete each model parameter corresponding to small loss change with its theoretical evidence of capturing sensitivity in DNNs. The loss change can be written as a form of Taylor expansion:
\begin{equation}
\label{eqn:losschange}
    \Delta\mathcal{L} \approx \underbrace{\frac{\partial\mathcal{L}}{\partial w}}_{\approx 0}\Delta w + \frac{1}{2}\Delta w^T \mathrm{H} \Delta w,
\end{equation}
where it satisfies $\Delta\mathcal{L} = \mathcal{L}(w+\Delta w)- \mathcal{L}(w)$, and $w$ denotes local \textit{Maximum A Posteriori} (MAP) parameters after training DNNs with its convergence. Note that there is an assumption that the first gradient of loss function closes to zero due to using the well-trained DNNs with their local MAP parameters around a local mode at $w$. Here, $\mathrm{H}$ indicates Hessian matrix computed by the second derivative of loss function $\mathcal{L}$ at model parameters $w$, such that it satisfies $\mathrm{H}=\nabla_{w}^2\mathcal{L}$. In calculating Hessian, Fisher information is employed to approximate it: $\nabla_{w}^2\mathcal{L} \simeq \E_{\mathcal{D}}[ \nabla_{w}\mathcal{L}^2]$. (see Appendix A)

However, due to the intractability of calculating full Hessian matrix $\mathrm{H}$ in DNNs, they have focused on its diagonal terms to approximate the loss change $\Delta\mathcal{L}$. The formulation for estimating the loss change can be written as follows:
\begin{equation}
\label{eqn:obdobs}
    \min\limits_{\Delta w \in \mathbb{R}^{d}}\frac{1}{2}\Delta w^T \mathrm{H} \Delta w, \quad\text{s.t. } e_{k}^{T}\Delta w + w_{k}=0,
\end{equation}
where $e_{k}$ denotes a canonical basis vector in $k^{\text{th}}$ dimension. With only of the diagonal considered, the corresponding solution for the loss change is deterministic to $\Delta\mathcal{L}_{\text{OBD}}=\frac{1}{2}w_{k}^2\mathrm{H}_{kk}$ where a single parameter $w_{k}$ is pruned. Once $\Delta\mathcal{L}_{\text{OBD}}$ is computed for all model parameters, then model parameters are cut out in order of small loss change by sorting $\Delta\mathcal{L}_{\text{OBD}}$. Note that $d$ specifies total size of model parameters $w$ in DNNs. In addition, Hessian is a symmetric matrix having shape of $\mathbb{R}^{d\times d}$ with positive semi-definite.

{\textbf{Optimal Brain Surgeon (OBS).}} OBD only addresses diagonal terms in Hessian matrix assuming model parameters are uncorrelated. Thereby, it is impossible to compute the loss change considering correlated model parameters in DNNs for generalization. To handle such limitation, Hassibi \etal~\cite{hassibi1993optimal} have proposed to compute the loss change with regards to full Hessian matrix. Its corresponding solution is also deterministic to $\Delta\mathcal{L}_{\text{OBS}}=\frac{1}{2}w_{k}^2/[\mathrm{H}^{-1}]_{kk}$ by Lagrangian relaxation (see Appendix B). After getting $\Delta\mathcal{L}_{\text{OBS}}$ for all model parameters, OBS equally removes model parameters in order of the small loss change.

\subsection{Masking Adversarial Damage (MAD)}

Although its successful achievements in standard pruning, it is difficult to directly apply it to adversarial pruning due to one major reason. The main principle of possibly computing the loss change $\Delta\mathcal{L}$ in \cref{eqn:losschange} is based on the fact that $\mathcal{L}(w)$ is an important criterion to ensure highly predictive DNNs. However, once adversarial examples are given, $\mathcal{L}(w)$ cannot be operated to be predictive anymore, where the first order of Taylor expansion $\frac{\partial\mathcal{L}}{\partial w}$ is no longer zero in adversarial settings. It is attributed to the performance degradation derived from the fragility of DNNs.

Hence, we need a highly optimized loss $\mathcal{L}(w)$ to construct \cref{eqn:losschange}, which signifies still predictive on the adversarial examples. Nonetheless, most of the existing adversarially trained models do not reach satisfactory robustness under adversarial attack scenario compared with their benign accuracy. To tackle such a problem, we employ adaptive masks directly multiplying to model parameters and optimize them to meet local convergence to be a highly predictive model for adversarial examples. The formulation of finding the masks can be written as follows:
\begin{equation}
\label{eqn:mask_opt}
    m^{*} =\arg\min\limits_{m\in\mathbb{R}^{d}}\max\limits_{\left \| \delta \right \|_{\infty} \leq \gamma}\mathcal{L}(f_{w_{m}}(x+\delta), y),
\end{equation}
where a model $f$ is parameterized by mask-applied parameters $w_{m}=w\odot m$. Here, $\odot$ indicates the Hadamard product. Note that dimension of model parameters $w$ is same as that of masks $m$, thus $d$ equally represents total size of $w$ and $m$ in DNNs. In addition, $m$ belong to continuous real number set in $[0, 1]^{d}\in\mathbb{R}^{d}$.

Through the mask optimization, we can effectively address the aforementioned issue to compute the loss change in adversarial settings. Then, we can re-interpret the loss change for the given adversarial examples as follows:
\begin{equation}
\label{eqn:mad}
    \mathcal{L}_{x_{\text{adv}}}(w) - \mathcal{L}_{x_{\text{adv}}}(w_{m^{*}}) = \frac{1}{2}\Delta w^T \mathrm{H} \Delta w,
\end{equation}
where $\mathcal{L}_{x_{\text{adv}}}(w)$ indicates an adversarial loss that have a high value due to broken model predictions for the given adversarial examples: $x_{\text{adv}}=x+\delta^{*}$. In addition, $\mathcal{L}_{x_{\text{adv}}}(w_{m^{*}})$ denotes a highly optimized adversarial loss against $x_{\text{adv}}$, which has a smaller value than previous one. Here, Hessian $\mathrm{H}\in\mathbb{R}^{d \times d}$ represents second-order information helping to build local geometry of the model parameters for capturing adversarially saliency factors.

Compared with previous works~\cite{lecun1990optimal, hassibi1993optimal} that remove a single parameter each, the procedures of \cref{eqn:mask_opt} and \cref{eqn:mad} enable to prune multiple parameters at once, which have a great advantage of considering impacts of intrinsic correlated connectivity among multiple parameters for model predictions. 
%of model prediction with multiple parameters correlated. 
On the other hand, to feasibly make use of the loss change, we should find out how to possibly calculate a parameter variation, denoted by $\Delta w$. In OBD and OBS, they have the tractable constraint for the variation $\Delta w$ removing a single model parameter: $e_{k}^{T}\Delta w + w_{k}=0$. Therefore, they can approach the loss change by a deterministic solution, as mentioned in \cref{sec:obd}.

However, we cannot straightforwardly get the deterministic solution for \cref{eqn:mad}, since the masks are tangled with the variation $\Delta w$ in DNNs. As long as such limitation exists, we should practically figure out $\Delta w$ with respect to the masks $m$. Thus, we provide the following constraint equation for pruning multiple parameters at once:
\begin{equation}
\label{eqn:constraint}
    w \odot (1-m) + \Delta w = \textbf{0},
\end{equation}
where $\textbf{0}$ indicates zero vector in $\mathbb{R}^{d}$. For extreme mask case, once $k^{\text{th}}$($=1,2,\cdots,d$) mask denoted by $m_{k}$ get closes to zero value such that $w_{m_{k}} = w_{k}m_{k} = 0$, then \cref{eqn:constraint} satisfies $w_{k} + \Delta w_{k} = 0$ which explicitly means that the $k^{\text{th}}$ mask-applied parameter $w_{m_{k}}$ is removed to zero.

Conversely, once $m_{k}$ gets close to one value such that $w_{m_{k}} = w_{k}$, then the constraint equation satisfies $\Delta w_{k} = 0$, which implies that $w_{m_{k}}$ is not a target to remove. If a mask is above zero value within one, it provokes an effect of smoothing a parameter variation $\Delta w$, which signifies that $w_{m_{k}}$ is not strictly erased yet some of them remains.

From these perspectives, masks can be formulated with parameter variation $\Delta w$, regarding correlation for all of the model parameters in DNNs. To compromise it on the variation $\Delta w$, we apply this constraint to \cref{eqn:mad}. Then, we can simply present the loss change in adversarial settings. At this point, we define the overall procedures of computing the loss change and pruning as \textit{Masking Adversarial Damage} (MAD). Here, the loss change of MAD $\Delta\mathcal{L}_{\text{MAD}}$ can be formulated as:
\begin{equation}
\label{eqn:mad_calculate}
    \Delta\mathcal{L}_{\text{MAD}} = \frac{1}{2}[w \odot (1-m^{*})]^T \mathrm{H} [w \odot (1-m^{*})],
\end{equation}
where it satisfies $\Delta\mathcal{L}_{\text{MAD}}=\mathcal{L}_{x_{\text{adv}}}(w) - \mathcal{L}_{x_{\text{adv}}}(w_{m^{*}})$. Here, we call the loss change $\Delta\mathcal{L}_{\text{MAD}}$ as \textit{adversarial saliency} to indicate its effect to model prediction. This mathematical expansion of MAD is completely aligned with theoretical analysis of Taylor expansion (see Appendix C).

Beyond its numerical loss change by itself, it provides us internal information in DNNs about which model parameter is a much or less adversarially important factor. Once we compute adversarial saliency for all model parameters and observe that some parameters have large saliency, we infer that they easily flip model predictions for adversarial examples, even with their small changes. Thus, they can be regarded as salient factors to get a knowledge of the robustness through adversarial training. On the other hand, in the contrary case, we interpret that the model parameters with small saliency are invariant to the model prediction. Hence, we can think of they are insignificant against adversarial examples. Accordingly, depending on the value size of adversarial saliency, we can identify which model parameters can be pruned without weakening adversarial robustness.

\subsection{Block-wise K-FAC Approximation}
In the previous section, we have focused on how to develop the loss change $\Delta\mathcal{L}_{\text{MAD}}$ in adversarial settings and how to compute variation $\Delta w$ despite the absence of a deterministic solution. However, there is another important issue to compute second-order information as Hessian. This is because Hessian is a huge matrix in terms of total parameter dimension, thus it remains challenging to efficiently reduce computational cost. To address it, there have been several methods of approximating Hessian by empirical Fisher Information~\cite{hassibi1993second, amari1998natural, graves2011practical, dong2017learning, zhang2018noisy, ritter2018scalable}, which can be written as:
\begin{equation}
    \label{eqn:fisher}
    \mathrm{H}\simeq F = \frac{1}{N}\sum_{n=1}^{N}\{\nabla_{w}\mathcal{L}_{n}\} \{\nabla_{w}\mathcal{L}_{n}\}^T,
\end{equation}
where $\nabla_{w}\mathcal{L}_{n}$ indicates $\nabla_{w}\mathcal{L}(f_{w}(x_n),y_n)$ in $n^{\text{th}}$ data sample such that $N$ represents the total number of data samples. $F$ denotes Fisher information matrix to possibly approximate Hessian in an empirical manner. Recently, various efficient approximation to Hessian has been proposed under \textit{Kronecker-Factored Approximate Curvature} (K-FAC)~\cite{martens2015optimizing, grosse2016kronecker, wang2019eigendamage} and WoodFisher approximation~\cite{singh2020woodfisher}.

However, since classical loss change (\cref{eqn:losschange}) for standard pruning has deterministic solutions for OBD and OBS, the several works of approximating Hessian does not have to fully compute the loss change. Instead, they figure out the loss change of $\Delta\mathcal{L}_{\text{OBD}}=\frac{1}{2}w_{k}^2\mathrm{H}_{kk}$ and $\Delta\mathcal{L}_{\text{OBS}}=\frac{1}{2}w_{k}^2/[\mathrm{H}^{-1}]_{kk}$, where diagonal elements of Hessian or inverse Hessian are only used. On the other hand, in order to acquire precise $\Delta\mathcal{L}_{\text{MAD}}$ (\ie, adversarial saliency), we should need to even consider off-diagonal terms in Hessian, but using full Hessian gives a striking computational burden. Hence, we introduce a block-wise K-FAC approximation to regard some blocks of the off-diagonal terms. The following Lemma \ref{lemma:1} describes K-FAC approximation, which efficiently approximates Hessian in DNNs with convolution layer. K-FAC for the fully-connected layer is described in Appendix D, where the differences are dimension scales of spatial size and its corresponding gradient.

\begin{lemma}
\label{lemma:1}
\textbf{Kronecker-Factored Approximate Curvature.} \\ At $l^{\text{th}}$ convolution layer in DNNs, let us define activation map $a\in\mathbb{R}^{c_{\text{in}}\times s_{a} }$, weight matrix $\mathcal{W}\in\mathbb{R}^{c_{\text{out}}\times c_{\text{in}}k^2}$, and layer output $z\in\mathbb{R}^{c_{\text{out}}\times s_{z}}$ such that it satisfies $z=\mathcal{W} \ast a$, $\nabla_{\mathcal{W}}\mathcal{L}=\sum_{i}\left[\nabla_{z_{i}}{\mathcal{L}} \times a^T_{i}\right]$, and $F=\E[\{\nabla_{\mathcal{W}}\mathcal{L}\} \{\nabla_{\mathcal{W}}\mathcal{L}\}^T]$, where $i$ denotes spatial index. Note that $c_{\text{in}}$ denotes channel number of $l^{\text{th}}$ layer, $c_{\text{out}}$ denotes channel number of $(l+1)^{\text{th}}$ layer, $s_{a}$ denotes spatial size of activation map $a$, $s_{z}$ denotes spatial size of layer output $z$, and $k$ denotes kernel size of weight.
\[
F \approx \sum_{i}\E[ \{\nabla_{z_i}\mathcal{L}\}\{\nabla_{z_i}\mathcal{L}\}^T] \otimes \sum_{i}\E[ a_ia_i^T] = \mathcal{Z} \otimes \mathcal{A},
\]
where $\mathcal{Z}\in\mathbb{R}^{c_{\text{out}}\times c_{\text{out}}}$ and $\mathcal{A}\in\mathbb{R}^{c_{\text{in}}k^2\times c_{\text{in}}k^2}$ stand for correlation of layer output $z$ and activation map $a$, respectively.
\end{lemma}

Decomposing Fisher information matrix $F$ into each Kronecker Factor for the correlations $\mathcal{Z}$ and $\mathcal{A}$ with relatively low dimension not only reduces storage cost but also enables efficient computation. Then, we first expand the adversarial saliency $\Delta \mathcal{L}_{\text{MAD}}$ using K-FAC, which can be written in $l^{\text{th}}$ convolution layer as:
\begin{equation}
    \label{eqn:k-fac mad}
    \begin{split}
    \Delta\mathcal{L}_{\text{MAD}} &= \frac{1}{2}\Delta w^T \mathcal{Z} \otimes \mathcal{A} \Delta w\\
    &=  \frac{1}{2} \text{Tr}\left[ \Delta w \Delta w^T \mathcal{Z} \otimes \mathcal{A}  \right],
    \end{split}
\end{equation}
where the variation is denoted by $\Delta w = -w \odot (1-m)$ for MAD constraint, of which vector shape is $\mathbb{R}^{d=c_{\text{out}}c_{\text{in}}k^2}$ in the convolution layer.

Here, we develop a block-wise K-FAC that allows for efficiently computing the adversarial saliency $\Delta \mathcal{L}_{\text{MAD}}$. Since we cannot use full Hessian for all model parameters due to computational limitation and physical memory budget, we instead deal with only of the diagonal terms in the correlation $\mathcal{Z}$ of layer output $z$, where we consider off-diagonal terms in $\mathcal{Z}$ as zero such that $\mathcal{Z}_{ij}=0$ $(i\neq j)$. Then, Fisher information matrix can be described with the block-wise K-FAC as follows:
\begin{equation}
    F\approx \mathcal{Z} \otimes \mathcal{A}=
    \label{eqn:block MAD}
    \begin{bmatrix}
    \mathcal{Z}_{11}\mathcal{A} & 0 & \cdots & 0 \\
    0 & \mathcal{Z}_{22}\mathcal{A} & 0 & \vdots  \\
    \vdots  & 0 & \ddots & 0 \\
    0 & \cdots & 0 & \mathcal{Z}_{rr}\mathcal{A} 
    \end{bmatrix},
\end{equation}
where $r=c_{\text{out}}$. By using this Fisher matrix $F$, we can efficiently compute the adversarial saliency from \cref{eqn:k-fac mad}, which can be formulated as follows (see Appendix E):
\begin{equation}
    \label{eqn:final_mad}
    \Delta\mathcal{L}_{\text{MAD}} = \sum_{i=1}^{r}\frac{\mathcal{Z}_{ii}}{2} \text{Tr}\left[ \Delta \mathcal{W}_{i} \Delta \mathcal{W}_{i}^T \mathcal{A}  \right],
\end{equation}
where we reshape variation vector $\Delta w$ to variation matrix $\Delta \mathcal{W}\in\mathbb{R}^{c_{\text{in}}k^2 \times c_{\text{out}}}$ such that $\Delta \mathcal{W} = -\mathcal{W} \odot (1-\mathcal{M})$, where mask matrix is also reshaped to $\mathcal{M}\in\mathbb{R}^{c_{\text{in}}k^2 \times c_{\text{out}} }$. Note that $\Delta \mathcal{W}_{i}$ denotes $i^{\text{th}}$ column vector in this variation matrix. In \cref{eqn:k-fac mad}, it has $\mathcal{O}(c_{\text{out}}^4 c_{\text{in}}^4 k^8)$ computation complexity due to full Hessian, but we can reduce it to $\mathcal{O}(c_{\text{out}}^2 c_{\text{in}}^4 k^8)$ by using the block-wise K-FAC. We get a computational benefit per convolution layer by $c_{\text{out}}^2(=r^2)$ as the matrix size of the correlation $\mathcal{Z}$ for layer output $z$.

\subsection{Pruning Low Adversarial Saliency}

Based on \cref{eqn:final_mad}, we finally get adversarial saliency for the given adversarial examples. In practice, there arises a natural phenomenon that the optimized mask $m^{*}$ overfits too much to specific samples of adversarial examples, thus it cannot be generalized to other adversarial examples. Therefore, after we optimize masks and compute the adversarial saliency for each data sample, we average them to generalize in all of the data samples. Then, by controlling pruning ratio $p\%$, we can readily prune model parameters by eliminating them in order of low adversarial saliency. Here, we describe \cref{alg:1} to explain MAD in detail.
%%%%%%%%%%%%%%%%%%%%%%%%%%%%%%%%%%%%%%%%%%%%%%%%%%%%%%%%%%%%%%%%%%%%
\begin{algorithm}[t!]
\setlength{\textfloatsep}{0pt} % Remove \textfloatsep
\caption{Masking Adversarial Damage}
\label{alg:1}
\begin{algorithmic}[1]
\Require pruning ratio $p\%$
\For {$(x, y)\sim\mathcal{D}$} \Comment{Finding Adversarial Saliency}
\State $x_{\text{adv}} \gets \text{attack}(x, y)$ \Comment{PGD Attack}
\State $m^{*} \gets \arg\min\limits_{m\in\mathbb{R}^{d}}\mathcal{L}(f_{w_{m}}(x_{\text{adv}}), y)$ \Comment{Finding Masks}

\For {$l^{\text{th}}$ layer index} 
\State $\Delta\mathcal{W}^{(l)}\gets - \mathcal{W}^{(l)} \odot (1-\mathcal{M}^{*(l)})$ \Comment{Constraint}
\For {$i^{\text{th}}$ channel index} \Comment{Block-wise K-FAC}
\State $\mathcal{B}_{ii}^{(l)}\gets\frac{\mathcal{Z}_{ii}^{(l)}}{2}\{\Delta \mathcal{W}_{i}^{(l)}\} \{\Delta \mathcal{W}_{i}^{(l)}\}^T \mathcal{A}^{(l)}$
\EndFor
\State $\Delta \mathcal{L}_{\text{MAD}}^{(l)}\gets\Delta \mathcal{L}_{\text{MAD}}^{(l)}+\frac{1}{N}\text{diag}(\mathcal{B}^{(l)})$ \Comment{Average}
\EndFor
\EndFor
\State Sorting $\Delta\mathcal{L}_{\text{MAD}}$ and Pruning $p\%$ parameters
\For {epoch} \Comment{Adversarial Training ($\alpha:lr$)}
\For {$(x, y)\sim\mathcal{D}$}
\State $x_{\text{adv}} \gets \text{attack}(x, y)$ \Comment{PGD Attack}
\State $w \gets w - {\alpha}\frac{\partial}{\partial w}\mathcal{L}(f_{w}(x_{\text{adv}}), y)$ \Comment{Update}
\EndFor
\EndFor
\end{algorithmic}
\end{algorithm}
%%%%%%%%%%%%%%%%%%%%%%%%%%%%%%%%%%%%%%%%%%%%%%%%%%%%%%%%%%%%%%%%%%%%

%------------------------------------------------------------------------
\section{Experiments}
\label{sec:Experiments}

\subsection{Implementation Details}
{\textbf{Datasets and Networks.}} We conduct exhaustive experiments on three datasets and three networks. For dataset, we use CIFAR-10~\cite{krizhevsky2009learning} and SVHN~\cite{netzer2011reading} which are well known as standard low dimensional datasets which equally have 32$\times$32 dimensional images with 10 classes. In addition, to demonstrate generalization of MAD in a larger dataset, we adopt Tiny-ImageNet~\cite{le2015tiny} with 64$\times$64 pixels, which is a small subset of ImageNet separated in 200 different classes. For the three datasets, we train the following three networks: VGG-16~\cite{vgg}, ResNet-18~\cite{resnet}, and WideResNet-28-10~\cite{wideresent}. 

{\textbf{Mask Optimization.}} First, we adversarially train the networks based on a standard adversarial training (AT)~\cite{madry2018towards}. After we complete to adversarially train the networks, we perform mask optimization in \cref{eqn:mask_opt} to find out adversarial saliency for model parameters. To do this, we make use of Adam~\cite{kingma2014adam} with a learning rate of $0.1$ and momentum of $0.9$ for each data sample. In addition, we use 20 iterations to optimize masks, which allows sufficient convergence to predict well for adversarial examples. Note that since we also observe that in a standard network, mask optimization does not work well for the given adversarial examples~\cite{kim2021distilling}, thus we cover the adversarially trained network only. 

{\textbf{Pruning by MAD.}} We prune model parameters in order of low adversarial saliency. In addition, we adversarially train the networks under a sparse network across a pruning ratio $p\%$. For the setting of adversarial training, we follow the conventional settings for $l_{\infty}$ perturbation budget $8/255$. In addition, we generate adversarial examples by PGD attack~\cite{madry2018towards} with random restarts, where we set attack steps to $10$ and set attack step size to $0.0069$ in training. Especially, since adversarially training Tiny-ImageNet is a computational burden, we employ fast adversarial training~\cite{Wong2020Fast} with FGSM attack~\cite{43405} where the step size is set to $1.25\times 8/255$. For training hyper-parameter, we set a learning rate of $0.1$ with SGD~\cite{ruder2016overview} in $60$ epochs, and we early stop~\cite{pmlrv119rice20a} to reduce overfitting harming adversarial robustness. In addition, we take a step scheduler to lower the learning rate by $0.1$ times on each $30$ and $45$ epoch, and we set the weight decay parameter to $2\times 10^{-4}$. 

{\textbf{Adversarial Attacks.}} To fairly validate adversarial robustness, we adopt three standard attacks: FGSM~\cite{43405}, PGD~\cite{madry2018towards}, CW$_{\infty}$~\cite{CW}, and two more recent advanced attacks with a budget-aware step size-free: AP (Auto-PGD) and parameter-free: AA (Auto-Attack) introduced by Francesco \etal~\cite{croce2020reliable}. In inference, PGD and AP have $30$ steps with random starts where PGD has $0.0023$ step size and AP has momentum coefficient $\rho=0.75$. In addition, we use CW$_{\infty}$ attack on $l_{\infty}$ perturbation budget $8/255$ by employing PGD gradient clamping with CW objective~\cite{CW} on $\kappa=0$.

%%%%%%%%%%%%%%%%%%%%%%%%%%%%%%%%%%%%%%%%%%%%%%%%%%%%%%%%%%%%%%%%%%%%
\begin{table*}[t!]
\centering
\renewcommand{\tabcolsep}{1.5mm}
\resizebox{0.95\linewidth}{!}{
\begin{tabular}{clcccccccccccccc}
\Xhline{3\arrayrulewidth}
\multicolumn{1}{l}{}      & \multicolumn{1}{c}{}  & \multicolumn{7}{c}{CIFAR-10}  & \multicolumn{7}{c}{SVHN} \\
\cmidrule(lr){3-9}\cmidrule(lr){10-16}
\multicolumn{1}{l}{}      & \multicolumn{1}{c}{Method}   & AT            & TRADES        & MART          & LWM  & ADMM          & HYDRA         & MAD           & AT            & TRADES        & MART          & LWM  & ADMM          & HYDRA         & MAD \\
\cmidrule(lr){2-2}\cmidrule(lr){3-5}\cmidrule(lr){6-9}\cmidrule(lr){10-12}\cmidrule(lr){13-16}
\multicolumn{1}{c}{Model} & \multicolumn{1}{c}{Sparsity} & \multicolumn{3}{c}{$0\%$}                       & \multicolumn{4}{c}{$90\%$}                             & \multicolumn{3}{c}{$0\%$}                       & \multicolumn{4}{c}{$90\%$}                             \\
\hline
\multirow{7}{*}{VGG}    & \textit{Params}                       & \multicolumn{3}{c}{14.7M}                     & \multicolumn{4}{c}{1.5M}                             & \multicolumn{3}{c}{14.7M}                     & \multicolumn{4}{c}{1.5M}                             \\
\cmidrule(lr){2-2} \cmidrule(lr){3-5} \cmidrule(lr){6-9} \cmidrule(lr){10-12} \cmidrule(lr){13-16}
& Clean                        & \textbf{81.3} & 80.2          & 80.2          & 76.0 & 79.0          & 76.4          & \textbf{81.4} & \textbf{93.2} & 91.8          & 91.7          & 88.4 & 92.4          & 87.9          & \textbf{92.8} \\
& FGSM                         & 55.2          & 54.3          & \textbf{57.9} & 50.3 & 52.6          & 50.4          & \textbf{57.0} & 67.1          & 66.5          & \textbf{68.4} & 62.4 & 65.2          & 61.2          & \textbf{68.2} \\
& PGD                          & 50.2          & 50.5          & \textbf{54.2} & 46.8 & 47.4          & 47.4          & \textbf{51.8} & 56.0          & 57.3          & \textbf{60.4} & 53.3 & 54.9          & 54.6          & \textbf{58.4} \\
& CW$_{\infty}$                           & 48.6          & 48.1          & \textbf{48.8} & 43.0 	     & 45.0          & 44.8          & \textbf{47.1} & 52.3          & \textbf{52.6} & 52.2          & 49.7 & 49.5          & 49.6          & \textbf{52.0} \\
& AP                         & 48.7          & 49.0          & \textbf{52.3} &  42.7      & 44.8          & 46.8          & \textbf{50.0} & 52.4          & 54.1          & \textbf{57.0} & 49.1 & 48.7          & 51.8          & \textbf{54.1} \\
& AA                           & 45.9          & 46.1          & \textbf{47.0} &  42.4    & 42.9          & 43.5          & \textbf{45.1} & 48.5          & \textbf{49.9} & 49.0          & 48.2 & 46.9          & 47.5          & \textbf{50.8} \\
\hline
\multirow{7}{*}{ResNet} & \textit{Params}                       & \multicolumn{3}{c}{11.2M}                     & \multicolumn{4}{c}{1.1M}                             & \multicolumn{3}{c}{11.2M}                     & \multicolumn{4}{c}{1.1M}                             \\
\cmidrule(lr){2-2} \cmidrule(lr){3-5} \cmidrule(lr){6-9} \cmidrule(lr){10-12} \cmidrule(lr){13-16}
& Clean                        & \textbf{84.2} & 82.4          & 83.3          & 79.4 & \textbf{83.4} & 80.9          & 82.7          & \textbf{93.7} & 93.1          & 92.6          & 90.4 & \textbf{93.8} & 90.5          & 93.3          \\
& FGSM                         & 58.1          & 57.9          & \textbf{60.3} & 55.0 & 56.7          & 56.4          & \textbf{58.4} & 75.3          & 81.3          & \textbf{81.7} & 68.9 & 72.0          & 70.6          & \textbf{74.4} \\
& PGD                          & 53.3          & 54.4          & \textbf{56.4} & 51.5 & 50.9          & 52.5          & \textbf{53.0} & 59.6          & 63.0          & \textbf{65.0} & 58.8 & 57.7          & 59.2          & \textbf{60.6} \\
& CW$_{\infty}$   &52.4          & 52.2          & \textbf{52.8} &  48.9 	    & 49.2          & 49.6  & \textbf{56.1} & 56.7          & \textbf{59.3} & 58.6          & 52.7 & 52.6          & 55.1          & \textbf{69.8} \\
& AP & 51.8          & 53.2          & \textbf{55.1} & 48.2    & 48.8          & \textbf{51.8} & 51.6  & 54.4          & 55.4          & \textbf{58.3} & 51.8 & 51.2          & 56.2 & \textbf{58.9}          \\
& AA  & 49.4          & 50.6          & \textbf{51.2} &    	47.8   & 46.9          & 47.6     & \textbf{48.2} & 50.5          & \textbf{49.4} & \textbf{49.4} &  51.6     & 48.7          & 51.0 & \textbf{53.2}          \\\hline

\multirow{7}{*}{WRN} & \textit{Params} & \multicolumn{3}{c}{36.5M}  & \multicolumn{4}{c}{3.7M}  & \multicolumn{3}{c}{36.5M} & \multicolumn{4}{c}{3.7M} \\
\cmidrule(lr){2-2} \cmidrule(lr){3-5} \cmidrule(lr){6-9} \cmidrule(lr){10-12} \cmidrule(lr){13-16}
& Clean &\textbf{87.6} & 86.9          & 87.2          & 85.6 & 87.4          & 87.4          &  \textbf{88.0 }        & \textbf{94.0} & 92.9          & 92.5          & 91.1 & \textbf{94.0}          & 91.0          & 93.8 \\

& FGSM   &61.7          & 61.6          & \textbf{63.6} & 60.3 & 59.8          & 61.6          &    \textbf{61.9}    & 73.6          & 74.0          & \textbf{76.9} & 68.9 & 70.6          & 69.0          & \textbf{74.9} \\

& PGD                          & 55.2          & 56.3          & \textbf{57.6} & 53.3 & 52.3          & 54.0          &  \textbf{55.2}     & 59.9          & 60.9          & \textbf{64.5} & 59.4 & 58.6          & 60.3    &\textbf{61.2} \\

& CW$_{\infty}$  &55.1       & \textbf{56.3} & 55.7          & 54.0     & 52.6          & \textbf{54.4}          &  \textbf{54.4} & 57.5          & 57.7          & \textbf{57.8} &   54.8   & 54.7          & 56.4          & \textbf{57.5} \\
& AP  & 53.9 & 55.1 & \textbf{55.5} &51.8 &52.1 &52.5 &\textbf{53.2} &56.0 &57.7          &\textbf{60.4} & 53.5     & 54.3          & 58.1  &\textbf{59.6}     \\
& AA &52.7  &\textbf{53.9} & 53.0          &  48.2    & 49.4          & 50.8          &\textbf{51.5}      & 52.7          & \textbf{54.1} & 53.3          &51.2     &52.0            & 54.4 & \textbf{55.1} \\
\Xhline{3\arrayrulewidth}
\end{tabular}
}
\vspace{-0.1cm}
\caption{Comparing adversarial robustness by pruning ratio $90\%$ (Sparsity) on low dimensional datasets: CIFAR-10 and SVHN trained with VGG-16, ResNet-18, and WideResNet-28-10. The bold expression denotes the best performance within the same sparsity. In addition, \textit{Params} indicates the number of non-zero model parameters.}
\vspace{-0.4cm}
\label{table:1}
\end{table*}
%%%%%%%%%%%%%%%%%%%%%%%%%%%%%%%%%%%%%%%%%%%%%%%%%%%%%%%%%%%%%%%%%%%%

\begin{table}[t!]
\centering
\renewcommand{\tabcolsep}{1.5mm}
\resizebox{0.9\linewidth}{!}{
\begin{tabular}{clcccccc}
\Xhline{3\arrayrulewidth} \rule{0pt}{9pt}
Model                   & \multicolumn{1}{l}{Method} & Clean         & FGSM          & PGD           & CW$_{\infty}$            & AP            & AA           \\
\hline
\multirow{3}{*}{VGG}    & LWM                        & 44.9          & 20.1          & 17.5          & 12.4          & 12.4          &  11.6             \\
                        & HYDRA                      & 46.1          & 20.6          & 17.8          & 12.8          & 14.8          & 11.8          \\
                        & MAD                        & \textbf{47.0} & \textbf{21.5} & \textbf{18.9} & \textbf{14.4} & \textbf{15.0} & \textbf{12.7} \\
                        \hline
\multirow{3}{*}{ResNet} & LWM                        & 47.4          & 22.7          & 20.6          & 15.4          & 15.3          & 13.8          \\
                        & HYDRA                      & 49.7          & 23.9          & 21.0          & 15.6          & 18.0          & 14.2          \\
                        & MAD                        & \textbf{50.7} & \textbf{25.4} & \textbf{21.9} & \textbf{17.3} & \textbf{18.5} & \textbf{15.6} \\
                        \hline
\multirow{3}{*}{WRN}   & LWM                        & 52.5          & 24.7          & 20.6          & 16.7          & 16.7          & 12.2          \\
                        & HYDRA                      & 53.5          & 24.8          & 21.0          & 16.4          & 17.3          & 14.5          \\
                        & MAD                        & \textbf{55.9} & \textbf{27.6} & \textbf{23.4}     & \textbf{19.8}     & \textbf{19.8}     & \textbf{17.2}  \\ 
\Xhline{3\arrayrulewidth} \rule{0pt}{9pt}
\end{tabular}
}
\vspace{-0.5cm}
\caption{Comparing adversarial robustness by pruning ratio $90\%$ (Sparsity) on a larger dataset: Tiny-ImageNet. All descriptions in this table are same as those of \cref{table:1}.}
\vspace{-0.5cm}
\label{table:2}
\end{table}

\subsection{Validating MAD}
{\textbf{Adversarial Robustness.}} To validate the effectiveness of MAD, we compare adversarial robustness with advanced recent defense methods: TRADES~\cite{pmlr-v97-zhang19p}, MART~\cite{Wang2020Improving} and strong baselines for adversarial pruning: LWM~\cite{sehwag2019towards}, ADMM~\cite{ye2019adversarial}, and HYDRA~\cite{sehwag2020hydra}. To fairly compare the performance, the defense methods and the baselines are aligned with our experiment setup, so that we build them equally based on adversarially trained model (AT) on PGD attack~\cite{madry2018towards}. \cref{table:1} demonstrates two perspectives: \romannum{1}) MAD mostly outperforms the robustness of the baselines. \romannum{2}) Despite using $10\%$ of model parameters, MAD shows less performance degradation of the robustness than the baselines compared by the best performance with full parameters. For all of the conducted attacks, MAD has $4.2(\%)$ and $1.9(\%)$ degradation for CIFAR-10 and SVHN each on average of three networks. For the baselines, HYDRA has $7.2(\%)$ and $6.1(\%)$ degradation, ADMM has $8.7$ and $8.3$, and LWM has $9.4$ and $8.4$ for CIFAR-10 and SVHN. Besides, for benign accuracy, MAD has $0.4(\%)$ and $0.4(\%)$ degradation, HYDRA has $3.4$ and $4.1$, ADMM has $1.4$ and $0.3$, and LWM has $4.8$ and $3.9$. Moreover, we also compare the robustness of MAD in Tiny-ImageNet with that of the baselines. As illustrated in \cref{table:2}, we corroborate that MAD also works well in a larger dataset, while preserving the robustness.

% For MAD, we have removed model parameters in order of low adversarial saliency. 

\begin{figure}[t!]
\centering
\includegraphics[width=0.99\linewidth]{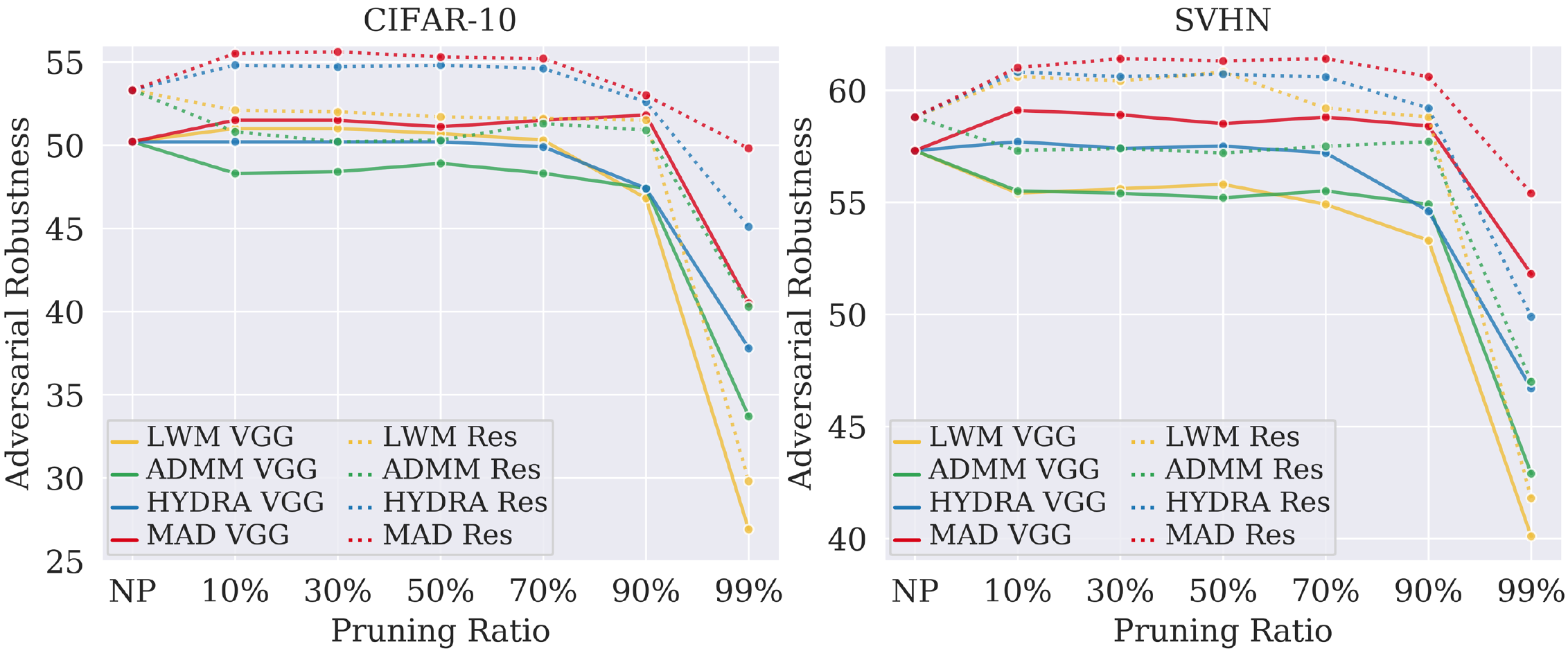}
\vspace{-0.2cm}
\caption{Ablation study results of adversarial robustness across varying pruning ratio. The experiments are conducted on CIFAR-10 and SVHN for VGG-16 and ResNet-18. Note that NP indicates AT with pruning ratio of $0\%$, ``Non-pruned".} 
\label{fig:pr}
\vspace{-0.5cm}
\end{figure}

{\textbf{Ablation Study.}} Despite its theoretical evidence of capturing adversarial saliency and its experimental improvements to the robustness, it is necessary to investigate that removing model parameters in reverse order induces much performance degradation. If it is correct, then we can exhibit adversarial saliency is operated well empirically in adversarial settings. \cref{table:3} shows the comparison of the robustness with three different cases. First one is randomly eliminating them (Random), and second is removing them in order of high adversarial saliency ($\Delta\mathcal{L}_{\text{MAD}}\uparrow$). Lastly, it is the case of MAD ($\Delta\mathcal{L}_{\text{MAD}}\downarrow$). The result in \cref{table:3} confirms that pruning model parameters in order of low adversarial saliency actually improves the adversarial robustness, and its reverse order pruning significantly loses the robustness. Furthermore, we wonder how MAD performs when model parameters are extremely pruned to $99\%$. Thereby, comparing MAD with the baselines, we validate the robustness across pruning ratio from $0\%$ (full parameters) to $99\%$ (extremely few parameters). According to \cref{fig:pr}, we observe that MAD shows better robustness than the baselines as the pruning ratio increases.

In Appendix F, we additionally experiment on Block-wise K-FAC to validate where the adversarial improvement of MAD derives from. Also, we discuss the violation of standard baselines (OBD/OBS) in adversarial settings.

 %%%%%%%%%%%%%%%%%%%%%%%%%%%%%%%%%%%%%%%%%%%%%%%%%%%%%%%%%%%%%%%%%%%%
\begin{table}[]
\centering
\renewcommand{\tabcolsep}{1.0mm}
\resizebox{0.9\linewidth}{!}{
\begin{tabular}{cccccccc}
\Xhline{3\arrayrulewidth} \rule{0pt}{9pt}
Sparsity             & Method               & Clean & FGSM & PGD  & CW$_{\infty}$ & AP   & AA   \\
\hline
\multirow{3}{*}{$90\%$} & Random               & 70.5  & 46.7 & 43.5 & 41.2          & 42.7 & 39.7 \\
                     & $\Delta \mathcal{L}_{\text{MAD}}\uparrow$   & 69.6  & 46.1 & 43.1 & 40.1          & 41.7 & 38.7 \\
                     \cmidrule{2-8}
                     & $\Delta \mathcal{L}_{\text{MAD}}\downarrow$ & \textbf{81.4}  & \textbf{57.0} & \textbf{51.8} & \textbf{47.1}          & \textbf{50.0} & \textbf{45.1} \\
                     \hline
\multirow{3}{*}{$99\%$} & Random               & 46.3  & 35.6 & 31.3 & 30.1          & 31.6 & 28.4 \\
                     & $\Delta \mathcal{L}_{\text{MAD}}\uparrow$   & 53.2  & 36.3 & 33.9 & 31.2          & 33.2 & 30.1 \\
                    \cmidrule{2-8}
                     & $\Delta \mathcal{L}_{\text{MAD}}\downarrow$ & \textbf{61.5}  & \textbf{43.0} & \textbf{41.0} & \textbf{60.0} & \textbf{39.8} & \textbf{35.0}\\
                     \Xhline{3\arrayrulewidth} \rule{0pt}{9pt}
\end{tabular}
}
\vspace{-0.5cm}
\caption{Ablation study results for the comparison of adversarial robustness in three different cases on pruning ratio $90\%$ and $99\%$. The validation is experimented on CIFAR-10 for VGG-16.}
\vspace{-0.4cm}
\label{table:3}
\end{table}
%%%%%%%%%%%%%%%%%%%%%%%%%%%%%%%%%%%%%%%%%%%%%%%%%%%%%%%%%%%%%%%%%%%%

 %%%%%%%%%%%%%%%%%%%%%%%%%%%%%%%%%%%%%%%%%%%%%%%%%%%%%%%%%%%%%%%%%%%%
\begin{figure*}[t!]
\centering
\includegraphics[width=0.9\linewidth]{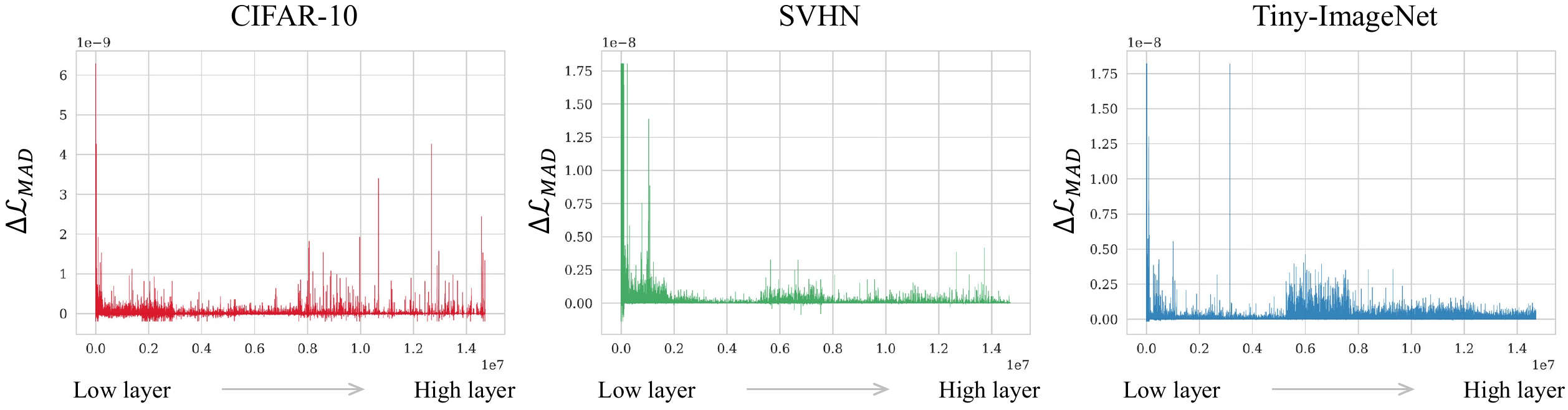}
\vspace{-0.3cm}
\caption{The shape of adversarial saliency to the whole model parameters in VGG-16 for CIFAR-10, SVHN, and Tiny-ImageNet.} 
\label{fig:as}
\vspace{-0.5cm}
\end{figure*}
%%%%%%%%%%%%%%%%%%%%%%%%%%%%%%%%%%%%%%%%%%%%%%%%%%%%%%%%%%%%%%%%%%%%

 %%%%%%%%%%%%%%%%%%%%%%%%%%%%%%%%%%%%%%%%%%%%%%%%%%%%%%%%%%%%%%%%%%%%
%%%%%%%%%%%%%%%%%%%%%%%%%%%%%%%%%%%%%%%%%%%%%%%%%%%%%%%%%%%%%%%%%%%%

\subsection{Analyzing Adversarial Saliency}
After we obtain an adversarially pruned model using MAD, what is our next question is \textit{“How is adversarial saliency distributed according to certain model parameters?”} To understand and track such sensitivity for adversarial examples, we plot adversarial saliency $\Delta\mathcal{L}_{\text{MAD}}$ from the model parameters of initial layer to the last layer in \cref{fig:as}. As in the figure, we observe the most salient parameters peak in the initial layers, sometimes referred as \textit{stem}. In standard training, it is well known that stem plays an important role to capture local features such as colours and edges, which will be integrated to recognize global objects~\cite{ramachandran2019stand}.

Based on the perspective of adversarial training considered as ultimate data  augmentation~\cite{tsipras2018robustness}, adversarial perturbation set is regarded as an invariant factor that the robust network should be independent of. From our pruning method, we observe that the model parameters of the stem layer are highly fragile in the parameter space, and they easily affect adversarial loss the most. Concurrently, it also indicates that training stem parameters invariant to the adversarial perturbation set is another decisive key of pruning approaches in adversarially trained networks and enhancing adversarial robustness.

\subsection{Understanding Compressed Features}
\begin{figure}[t!]
\centering
\includegraphics[width=0.9\linewidth]{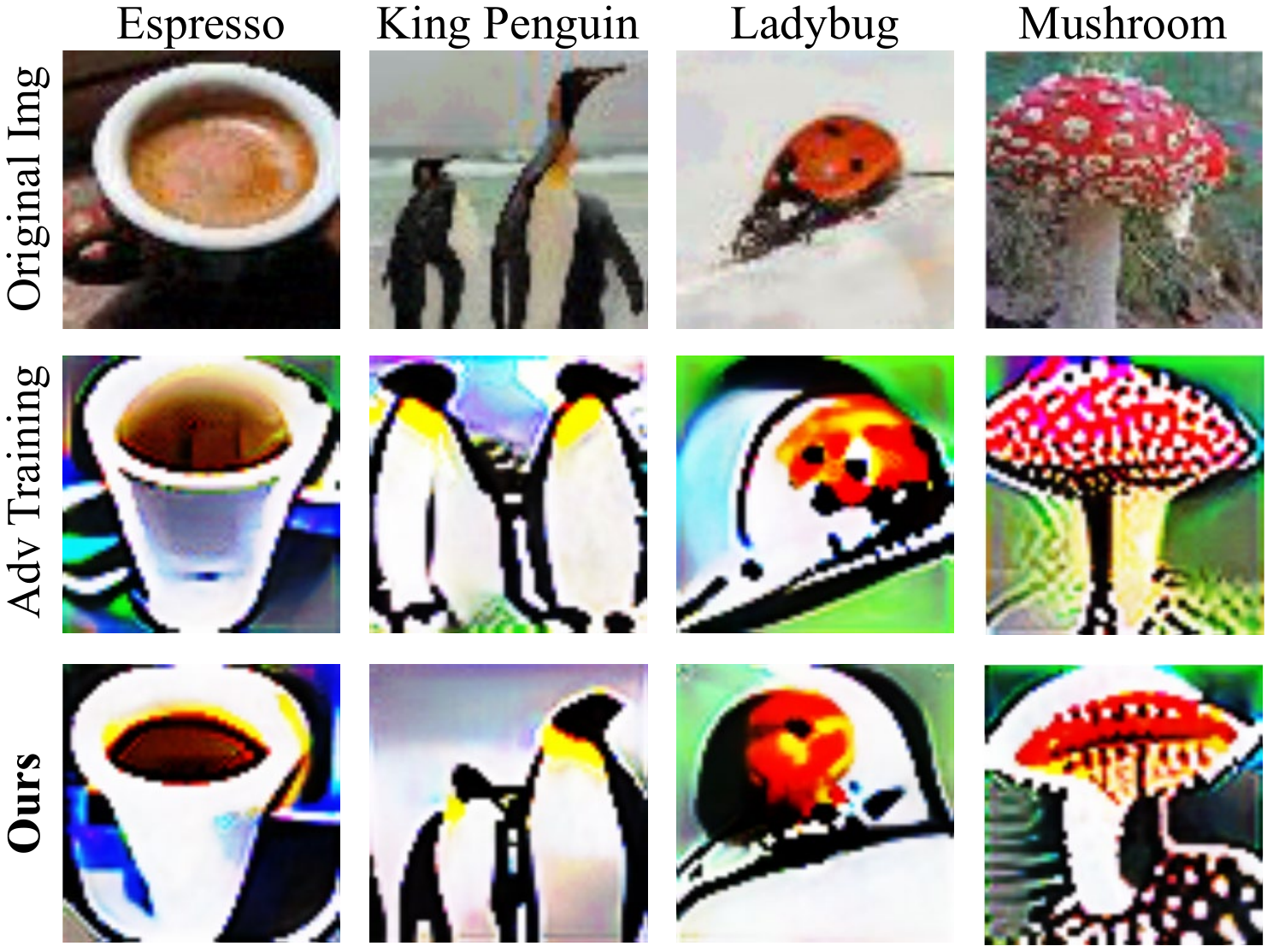}
\vspace{-0.3cm}
\caption{Feature visualization results of Tiny-ImageNet that include semantic information for target objects. We compare MAD with non-pruned model (AT) for VGG-16. We set the pruning ratio of MAD as $90\%$ and illustrate successfully defended cases.}
\label{fig:vs}
\vspace{-0.5cm}
\end{figure}

We have demonstrated the effectiveness of MAD in the model compression while retaining adversarial robustness. Then, we want to know if the compressed features can be identified in the intermediate feature space. Recently, several works~\cite{engstrom2019adversarial, kim2021distilling} have introduced a way of understanding adversarial feature representation using direct feature visualization methods~\cite{olah2017feature, nguyen2019understanding}. We adopt Olah \etal~\cite{olah2017feature} to analyze the feature representation of our pruned model and compare them with non-pruned model (AT) in \cref{fig:vs}.

As seen in this figure, we have observed that MAD can retain the semantic information of the intermediate feature representation, even with the high pruning ratio. Compared with the visualization results of the non-pruned model, MAD seemingly loses some feature representation, but still shows the recognizable quality in the semantic information of target objects regardless of the high pruning ratio.

 %%%%%%%%%%%%%%%%%%%%%%%%%%%%%%%%%%%%%%%%%%%%%%%%%%%%%%%%%%%%%%%%%%%%
%%%%%%%%%%%%%%%%%%%%%%%%%%%%%%%%%%%%%%%%%%%%%%%%%%%%%%%%%%%%%%%%%%%%

%------------------------------------------------------------------------
\section{Discussion and Conclusion}
\label{sec:Discussion}
% 필수적으로 Negative societal Impacts and Limitation
{\textbf{Discussion.}} Adversarial attacks can potentially cause negative impacts on various DNN applications due to high computation and its fragility. By pruning model parameters without weakening adversarial robustness, our work contributes important societal impacts in this research area. Furthermore, in our promising observation that model parameters of initial layer are highly sensitive to adversarial loss, we hope to progress in another future direction of utilizing such property to enhance adversarial robustness.

{\textbf{Conclusion.}} To achieve adversarial robustness and model compression concurrently, we propose a novel adversarial pruning method, Masking Adversarial Damage (MAD). By exploiting second-order information with mask optimization and Block-wise K-FAC, we can precisely estimate adversarial saliency of network parameters. Through extensive validations, we corroborate that pruning network parameters in order of low adversarial saliency retains adversarial robustness while alleviating performance degradation compared with previous adversarial pruning methods.

%------------------------------------------------------------------------
{
\hfill \break
\noindent{\textbf{Acknowledgments.}} This work was conducted by Center for Applied Research in Artificial Intelligence (CARAI) grant funded by DAPA and ADD (UD190031RD).
}

%%%%%%%%% REFERENCES START
\newpage
{\small
\bibliographystyle{ieee_fullname}
\bibliography{reference}
}
%%%%%%%%% REFERENCES END

\end{document}